\documentclass{article}
\usepackage{spconf,amsmath,graphicx,amssymb,multirow,multicol,booktabs}
\usepackage{hyperref}
\usepackage{xcolor}

\usepackage[small,compact]{titlesec} 
\usepackage{times}
\usepackage[small,it]{caption}

\title{Training ASR models by Generation of Contextual Information}
\name{\begin{tabular}{c}Kritika Singh, Dmytro Okhonko, Jun Liu, Yongqiang Wang, Frank Zhang, Ross Girshick, Sergey Edunov,\\ Fuchun Peng, Yatharth Saraf, Geoffrey Zweig, Abdelrahman Mohamed\end{tabular}}
\address{Facebook AI}

\begin{document}
%
\maketitle
\begin{abstract}
Supervised ASR models have reached unprecedented levels of accuracy, thanks in part to ever-increasing amounts of labelled training data. However, in many applications and locales, only moderate amounts of data are available, which has led to a surge in semi- and weakly-supervised learning research. In this paper, we conduct a large-scale study evaluating the effectiveness of weakly-supervised learning for speech recognition by using loosely related contextual information as a surrogate for ground-truth labels. For weakly supervised training, we use 50k hours of public English social media videos along with their respective titles and post text to train an encoder-decoder transformer model. Our best encoder-decoder models achieve an average of 20.8\% WER reduction over a 1000 hours supervised baseline, and an average of 13.4\% WER reduction when using only the weakly supervised encoder for CTC fine-tuning. Our results show that our setup for weak supervision improved both the encoder acoustic representations as well as the decoder language generation abilities. 
\end{abstract}
\begin{keywords}
End-to-end ASR, Weak-supervision
\end{keywords}
\section{Introduction}
Over the past few years, Automatic Speech Recognition (ASR) has made great strides due to the successful application of supervised Deep Learning (DL) techniques \cite{hinton_2012, parity_msr, parity_ibm, ds2}. However, one drawback of such approaches is the heavy reliance on large volume of supervision which can be difficult to acquire for new domains. A good ASR system operating in real environments requires a large volume of training data to marginalize out and deal with different acoustic conditions of background noise, languages, accents, speakers, and their emotional states. This practical need has led to a surge in ASR research in unsupervised acoustic feature learning \cite{pg_08, aren_10, glass_12, JHU_2012, JSALT_2017, s2v, cpc, w2v}, as well as semi and weakly-supervised learning \cite{Vesely_13, Sheng_2017, hari_2019, tara_sslearning, Alishahi_2017, ttic_may2017}. In \cite{ioc_13},  the ``island of confidence'' technique was used to filter the owner-uploaded video transcripts creating additional weakly-supervised ASR training data. $1$ million hours of audio were transcribed using a teacher ASR model to train a production-ready student model \cite{hari_2019}. To improve performance in rare words and proper nouns, \cite{tara_sslearning} distilled top hypothesis generated by a contextually-biased ASR system as ground truth for training an encoder-decoder ASR model.\\
This paper belongs to this last category with a focus on public social media videos, which provide interesting challenges and opportunities for ASR research. On one hand, these videos contain a diverse range of speakers, dialects, topics, and acoustic conditions making automatic recognition difficult. On the other hand, parallel audio, visual and text information (e.g. video title, post text, and comments) is available for social media videos over which joint multi-modal learning is possible. This work focuses solely on utilizing video title and post text as additional contextual information for acoustic model training.\\
The relationship between contextual text and audio associated with a video ranges from weak semantic relatedness to, sometimes, overlap of exact words, phrases, or quotes taken verbatim from the audio. Training an ASR model to generate such related context information from audio signals exposes it to a large volume of diverse training examples, even if they are far from the exact speech content of audio. The downside is that the audio content may not be related or represented at all in the contextual text, let alone being monotonically aligned to the audio content. In this study, we evaluate the effectiveness of using contextual text from videos as weak labels for large-scale ASR training, and outline a proposal to overcome the aforementioned problems, achieving an average of 20.8\% WER reduction over an encoder-decoder baseline system trained only on 1000h of supervised data, and 13.4\% when we transfer only the encoder part of the model to be fine-tuned using the Connectionist Temporal Classification (CTC) loss \cite{ctc}. 


\section{Weakly supervised training}
\subsection{Datasets}
We use two sets of training data: (i) $\{X, Y^s\} \in \mathcal{D}^s$ is the supervised data where $X$ and $Y^s$ are pairs of audio features and label sequences. (ii) $\{X, Y^w\} \in \mathcal{D}^w$ is the weakly-supervised dataset where $X$ and $Y^w$ are pairs of audio features and the corresponding contextual text. The targets $Y^s$ and $Y^w$ are sequences of sub-word units \cite{sp}.
\subsection{The proposed approach}
Our proposed acoustic model training centers around utilizing distant, weak supervision from contextual text surrounding social media videos. 
We use an encoder-decoder approach \cite{las, s2s_speech_MO} for maximizing the conditional probability of generating the contextual text sequence $Y^w$ given $X$ an input sequence of mel-scale log filterbank features where $x_i \in R^d$
\begin{equation*}
\begin{split}
    \mathcal{F}^w &= p(Y^w|X; \theta^w) \\
    &=\prod_{i=1}^M p(y_i^w | y_1^w, y_2^w, ..., y_{i-1}^w, x_1, x_2, ..., x_T; \theta^w)
\end{split}
\end{equation*}
The attention-based encoder-decoder approach fits well with the proposed training approach since it offers flexible alignment and unconstrained coverage between input and output sequences. Other ASR training approaches aren't suitable given the abstractive relationship between $Y^w$ and $X$. The hybrid HMM-NN approach requires a low-level sub-second alignment between input audio and output targets, while the CTC approach assumes a monotonic alignment between inputs and outputs, and it constrains the maximum possible length of the output sequence by the length of the input sequence.
The final objective function is $\mathcal{F} = \mathcal{F}^w + \mathcal{F}^s$ where the supervised term, $\mathcal{F}^s = p(Y^s|X; \theta^s)$, is also maximized using the encoder-decoder approach. We share the full model for both types of data, where $\theta^w = \theta^s = \{\theta_{enc}, \theta_{dec}\}$ combines the parameters in the audio encoder and the language generation and attention parameters in the decoder. During training, we alternate, with some mixing ratio, between mini-batches sampled from the two training sets $\mathcal{D}^s$ and $\mathcal{D}^w$.

\subsection{The main assumptions} 
We are making two main assumptions in the proposed training approach:\\ 
(1) $|\mathcal{D}^w| >> |\mathcal{D}^s|$, and the diversity of acoustic conditions represented in $\mathcal{D}^w$ is much larger than that in the supervised data $\mathcal{D}^s$. Therefore, training on $\mathcal{D}^w$ has the potential to improve the final model's ability to generalize better to new speakers and recording conditions compared to a baseline model trained only on $\mathcal{D}^s$. To test the importance of this assumption, in our experiments, we present results for pre-training the ASR model using 50x, 10x, and 2x of the supervised data size.\\
(2) Maximizing $p(Y^w|X; \theta)$ can be used as a proxy for maximizing $p(Y^s|X, \theta)$. This is a rather strong assumption since the best ASR system will not generate a commentary for its speech inputs and vice versa. In other words, $\theta^{w*}$, the optimal model parameters maximizing the conditional likelihood of $\mathcal{D}^w$, may not equal an optimal ASR model's parameters $\theta^{s*}$. To test this assumption, we explore three specific questions: \\
(i) Given an input sequence $X$, how close is the user-generated commentary $Y^w$ to $Y^s$, the true audio content, under some semantic measure of relatedness? We use the set intersection of words in $Y^w$ and hypothesis generated using a baseline ASR as a proxy for relatedness. In our experiments, we test multiple levels of strictness for enforcing this condition.\\ (ii) Does maximizing $p(Y^w|X; \theta)$ improve $p(Y^s|X, \theta)$ during all phases of model optimization? We distinguish between three learning phases during model optimization: (a) An initial \texttt{burn-in} phase (b) A final \texttt{fine-tune} phase (c) An intermediate \texttt{train-main} phase. More details about these three phases are in Section 2.5. We hypothesize that the maximum transfer between $\mathcal{F}^w$ and $\mathcal{F}^s$ happens during the \texttt{train-main} intermediate phase of learning.\\
(iii) Does training the ASR model on $\mathcal{D}^w$ benefit all model components equally? Does it hurt some of them? To answer this question, we evaluate the impact of weakly supervised training on two ASR supervised fine-tuning setups: (a) One that utilizes both $\theta_{enc}$ and $\theta_{dec}$ from weak-supervision i.e, the acoustic and the language modeling components, for the final ASR model fine-tuning. (b) A second ASR setup where we only use $\theta_{enc}$ for initializing an acoustic-only model that is subsequently fine-tuned with a CTC loss function \cite{ctc} using an independently trained and fixed language modeling component. Using these two setups, we can distinguish gains due to better encoder acoustic representation from better language generation abilities learned by the decoder component.

\subsection{The model architecture}
\begin{figure}
  \centerline{\includegraphics[width=8.5cm]{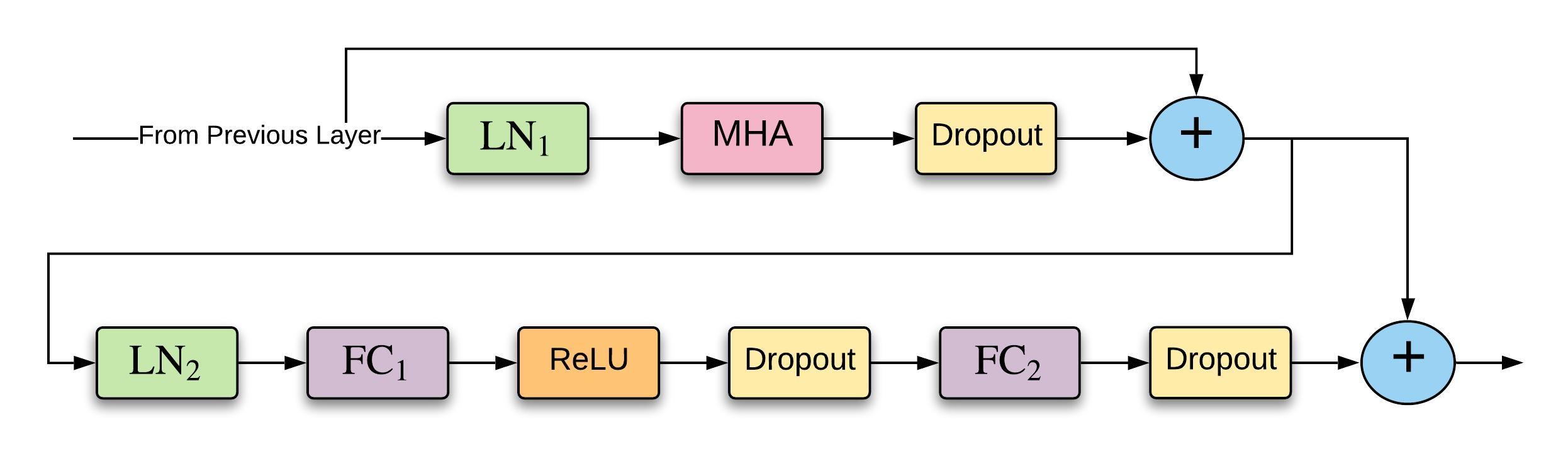}}
\caption{A block diagram of one transformer block.\vspace{1em}}
\label{fig:transformer}
\end{figure}

We use transformer blocks as building blocks in our encoder-decoder ASR model \cite{transformer, speech_transformer, trans_vs_rnn}, shown in figure \ref{fig:transformer}, and follow the convolutional transformer architecture from \cite{conv_trans}. For joint encoding of input content and position, the encoder input applies convolutional blocks each consisting of 2-D convolution, Layer Normalization (LN), ReLU non-linearity, and max-pooling layers:
\begin{equation*}
    [c_1, c_2, ..., c_\tau] = 2DConvBlocks([x_1, x_2, ..., x_T])
\end{equation*}
The encoder then applies multiple transformer blocks. Multi-headed self-attention (MHA) is the core component of transformer blocks. Self-attention (SA) represents each time step of the sequence $C \in \mathbb{R}^{\tau\times d}$ as a sum of all other time steps weighted by the inner products of their representations where each time step acts as a query $q_t$, a key $k_t$, and a value $v_t$. Scores for each time step are scaled by the inverse square root of the dimension $d$ over which the inner product is computed. A softmax operation is applied over all possible key indices, to encourage soft competition between different time steps, followed by a dropout operation to the combination weights: 
\begin{equation*}
    SA(Q,K,V) = Dropout\left(Softmax(QK^T/\sqrt{d})\right) * V
\end{equation*}
Multi-head attention (MHA) extends self-attention by repeating it number-of-heads times, $H$, using different linear projections for each head, and concatenating their outputs:
\begin{equation*}
    MHA(Q, K, V) = Concat_{h=1}^H\left( SA(Qw_q^h , Kw_k^h , Vw_v^h)\right) * w_o
\end{equation*}

$\{w_q^h, w_k^h, w_v^h\} \in \mathbb{R}^{d\times d_i}$ project Q, K, and V matrices differently for each head to the desired inner product dimension $d_i$. $w_o \in \mathbb{R}^{Hd_i\times d}$ projects the concatenated self-attention vectors to the output dimension $d$. Following the MHA sub-block, each transformer block applies a fully connected feed-forward sub-block, which is composed of two linear transformers and a ReLU non-linearity in between, to each time step. To avoid vanishing gradients, residual connections are added around the MHA and the fully connected sub-blocks, and LN operations are applied before them. The only difference between the encoder and decoder architectures is the use of a 1-d convolution operation to represent previously generated output tokens $[y_1, y_2, ..., y_{i-1}]$. The decoder component uses two transformer blocks each with multi-head cross attention for summarizing the final encoder representations. The dot-product cross-attention makes neither coverage nor monotonicity assumptions about the relationship between the input and output sequences. Similar to story and dialogue response generation \cite{angela_story_2018, msr_dialogue_2016}, segments of the input sequence can be covered in the output sequence once, multiple times or none at all, and similarly for segments of the output sequence.

\subsection{The training process} 
In our experiments, we study the impact of the three training phases introduced in 2.3: (1) An initial supervised \texttt{burn-in} phase in which the decoder cross-attention learns to properly communicate gradient information to adjust encoder acoustic features. (2) A training phase driven by a mixture of the supervised and the weakly supervised loss functions, we refer to it as the \texttt{train-main} phase, in which the model expands its inventory of audio features and mappings between acoustic and linguistic cues. (3) A final supervised-only \texttt{fine-tune} phase which utilizes either the full encoder-decoder model trained in the \texttt{train-main} step, or the encoder component to be refined by the CTC loss. Given the transformer's ability to reconstruct input sequences in any desired arbitrary order, an extra transformer block is optionally added to the encoder layers before fine-tuning to smooth out the transition from the encoder-decoder cross-entropy loss to the CTC loss which enforces monotonicity between input and output sequence.


\section{Experiments}
\textbf{Data:} Both the supervised and weakly-supervised (WS) datasets used in this study are sampled from our in-house datasets. Our supervised dataset consists of 1000 hours of data sampled from public English videos that are anonymized. We use this data exclusively in the \texttt{burn-in} and \texttt{fine-tune} stages of training, and on a fraction of minibatches in \texttt{train-main} determined by the mixing ratio, as well as for training the baseline ASR model.\\
Our weakly-supervised dataset consists of 4M public English videos that are 30 to 60 seconds in duration with contextual text between 60 to 700 characters, totaling 50,000 hours. We restrict our target context to video titles and post text only. Because the relevance of the contextual text to the audio content might greatly impact the expected gain, we filter the weakly supervised data based on the set intersection between contextual text and the baseline ASR hypothesis considering only words of length more than 3 characters, to create two additional sets: (1) A 2,300 hours subset with an intersection of 14 words or more (2) A 12,800 hours subset with an intersection of 6 words or more. To test the impact of quality of contextual text while keeping acoustic richness same, we create two additional subsets: (3) A 2,300 hours subset randomly sampled from the 50,000 hour original set (4) A 12,800 hours subset randomly sampled from the 50,000 hour original set. Additionally, to measure the impact of contextual information for weak supervision in terms of labeled data size, we create: (5) A 2,000 hours set of supervised data that is disjoint to the 1000 hours set used for the baseline. For performance evaluation, we use three test sets, \texttt{clean} with 1.3K videos (20 hours), \texttt{noisy} with 1.3K videos (20 hours), and \texttt{extreme} with 13K videos (77 hours) which is more acoustically and phonetically challenging. For hyper-parameter tuning and model selection, we use a \texttt{dev-noisy} subset which consists of 600 videos (9 hours). For inference using the encoder-only model, we use a 5-gram language model which is estimated from 1M utterances containing about 120k distinct tokens.\\
\textbf{Experimental setup:} The input speech is encoded into 80 dimensions of mel-scale log filterbank features computed over 16ms and shift of 10ms. The encoder two 2-D convolution blocks uses kernel size=3 and output features of 64 and 128 for each block respectively. The max-pooling layers sub-sample the input time steps and frequency channels by a factor of $4$. All encoder and decoder transformer blocks, 10 and 2 blocks respectively, use 1k hidden dimension, 16 heads, 4k projection layer before the ReLU nonlinearity, and dropout rate of 0.15. The decoder part uses 4 1-D convolutional layers with kernel size=3 and output features of 256. Supervised labels and contextual text is encoded into 5k sub-word output vocabulary \cite{sp}. We use the AdaDelta algorithm \cite{adadelta} with fixed learning rate=1.0 and gradient clipping at 10.0 where total gradients are scaled by the number of utterances in each minibatch. During \texttt{train-main} we save checkpoints every 5k model updates and average the last 20 checkpoints to initialize the \texttt{fine-tune} phase. We also average the last 20 checkpoints of fine-tuning before decoding. We use a beam size=20 for encoder-decoder model inference without any external language model. We use the 5-gram LM for decoding the CTC fine-tuned models.
\begin{table}
\centering
\setlength\tabcolsep{4.0pt}
 \begin{tabular}{cccccccc} 
\toprule
 \multirow{2}{*}{} &\multicolumn{3}{c}{Encoder-Decoder}&& \multicolumn{3}{c}{CTC} \\ 
 \cline{2-4}\cline{6-8}
 & clean & noisy & extreme && clean & noisy & extreme \\
 \midrule
 \multicolumn{8}{c}{Supervised baseline}\\
 \midrule
 1000h & 22.8 & 30.2 & 42.1 && 21.6 & 28.5 & 37.6 \\ 
 \midrule
\multicolumn{8}{c}{Weakly supervised models}\\
\midrule
 2300h & 20.9 & 27.5 & 38.2 && 19.2 & 25.8 & 35.2\\
 12,800h & 18.6 & 25.5 & 34.8 && 18.7 & 25.2 & 34.2\\
 50,000h & 18.3 & 25 & 34.6 && 18.6 & \textbf{24.5} & 34.2\\
\midrule
\multicolumn{8}{c}{Weakly supervised filtered by relevance}\\
\midrule
 2300h & 19.3 & 26.3 & 37 && 18.7 & 25.2 & 34.6\\
 12,800h & \textbf{17.6} & \textbf{24.3} & \textbf{33.7} && \textbf{18.2} & 24.8 & \textbf{33.7}\\
 \midrule
\multicolumn{8}{c}{Extra true labels instead of weak supervision}\\
\midrule
 2000h & 18.8 & 25.8 & 35 && 18.7 & 25.1 & 34\\
\bottomrule
\end{tabular}
\caption{WERs of the enc-dec and CTC fine-tuned models on the test sets \texttt{clean}, \texttt{noisy} and \texttt{extreme} for different train data sizes}
\label{tab1}
\end{table}
\\\textbf{Results:} Table \ref{tab1} presents the main results of this study on the three test conditions showing an average WER reduction of 20.8\% for the encoder-decoder setup and 13.4\% for the CTC setup compared to the baseline supervised model. For all experiments shown in table \ref{tab1}, the \texttt{burn-in} phase had 15k model updates, \texttt{train-main} had 400k updates with almost one third of the minibatches sampled from the supervised data (mixing ratio=0.3). Supervised \texttt{fine-tune} phase used 22k model updates for the encoder-decoder architectures and 150k updates for the encoder-only CTC loss.

While using the full 50,000h weakly supervised data improves upon the baseline system, filtering it for relevant content provided the best performance almost across all setups. Both models show improvement from using more weakly supervised data, however the encoder-decoder model's gains are slightly higher given that large weak supervision data improves both their acoustic encoder representations and decoder generation abilities, relative to the CTC fine-tuned model which uses a fixed n-gram language model and benefits only from the improved encoder representations.

An interesting observation is that, using weak supervision, the encoder-decoder setups are almost as good or slightly better than their corresponding encoder-only setups, even though encoder-decoder models don't use any external language model. This suggests that weak supervision via context generation was enough to realize a language modeling capability similar to that of the n-gram LM used for the encoder-only models. The best weakly supervised systems are consistently better than using an additional 2000 hours of labeled data for the \texttt{train-main} phase. This magnifies the value of weak supervision on reducing the requirement for ASR data labeling. 
\begin{table}
\centering
\setlength\tabcolsep{9.0pt}
 \begin{tabular}{cccccc} 
 \toprule
 Burn-in phase && \multicolumn{2}{c}{N} & \multicolumn{2}{c}{Y} \\
\cline{1-1}\cline{3-6}
 Mixing ratio && 0 & 0.3 & 0 & 0.3 \\
 \midrule
 50,000h && 27.6 & 27.3 & 24.7 & 25.4\\
 Filtered 12,800h && 28.7 & 26.1 & 24.4 & 25\\
 
\bottomrule
\end{tabular}
\caption{WERs of fine-tuned CTC model on \texttt{dev-noisy} under different conditions of burn-in and mixing ratio}
\label{tab2}
\vspace{-6mm}
\end{table}
One problem of using weak supervision that is not aligned with the input sequence is that the decoder won't be able to refine encoder representations easily. Hence we included supervised mixing and/or initial burn-in phase during our weakly supervised \texttt{train-main} phase. Table \ref{tab2} shows the effect of both techniques on the learned representations of the encoder-only setup trained by the CTC loss. Comparing cases either with burn-in or mixing shows that mixing helps a bit, however, supervised \texttt{burn-in} is much more important than mixing for encoder representations. When burn-in is on, spending almost one third of the mini-batches during \texttt{train-main} visiting supervised data seems to hurt performance because the model has less chance to observe the more diverse and larger weakly supervised data.
\begin{table}
\centering
\setlength\tabcolsep{10.0pt}
 \begin{tabular}{ccc} 
 \toprule
 Enc-Dec fine-tuning & 50,000h & Filtered 12,800h \\
\midrule
 N & 26.4 & 25.3\\
 Y & 25.7 & 24.8 \\
 \bottomrule
\end{tabular}
\caption{The impact of fine-tuning encoder-decoder model on \texttt{dev-noisy}}
\label{tab3}
\vspace{-6mm}
\end{table}
Table \ref{tab3} shows that the encoder-decoder model is ready for recognition with good performance even without the final \texttt{fine-tune} phase, but it still benefits from the 22k updates of supervised fine-tuning which, we believe, polish its decoder cross-attention input audio sequences to be more monotonic. 

\section{Discussion and Related Work}
Our work builds on the success of sequence-to-sequence learning for ASR, both the CTC-based \cite{e2e_asr, w2l, ds2} and the attention-based \cite{s2s_speech_MO, las} variants, by replacing the ground-truth target sequences with semantically related weak supervision.
This study is primarily motivated by \cite{uru, flicker} where hashtag prediction of social media images is successfully used for large CNN pre-training for many image classification and object detection tasks. There is a growing body of research in self-supervised pre-training of models on surrogate tasks for general language representation which have shown great success in several downstream NLP tasks \cite{bert, roBERTa}. Grounding language learning and generation \cite{Thiessen_10, gab_14, glass_16, Alishahi_2017, rahma_2017}, spoken keyword spotting, and audio representation \cite{ttic_may2017, Aytar_2016} on visual cues motivated this work where, similar to this study, inputs and outputs are loosely related with no guarantees of coverage. Weak semantic labels showed significant improvements in phonetic learning in a model of infant language acquisition and vocal commands learning for dysarthric speakers \cite{Stella_14, hugo_15}. This work belongs to the growing line of research focusing on reducing the reliance on supervised labels for building ASR systems through unsupervised unit discovery and acoustic representation learning \cite{pg_08, aren_10, glass_12, JHU_2012, JSALT_2017, s2v, cpc, w2v}, multi- and cross-lingual transfer learning in low-resource conditions \cite{Haihua_2016, jia_2015, Georg_2013, arnab_2013, Huang_2013, Ngoc_2014}, and semi-supervised learning \cite{Vesely_13, Sheng_2017, hari_2019, tara_sslearning}. 

\section{Conclusion and Future work}
We presented a large-scale weakly supervised training method for speech recognition systems that uses contextual social media information -- titles and post text -- as a surrogate for transcriptions. An encoder-decoder approach was trained to generate the contextual labels, which are semantically related to the spoken audio content, but have neither monotonicity nor full coverage. Our best models achieved averages of 20.8\% and 13.4\% WER reduction over supervised baselines. In the future, we would like to combine compare and investigate synergies between semi-supervised student-teacher modeling with our weak-supervision method.

\footnotesize
\bibliographystyle{IEEEbib}
\bibliography{refs}

\end{document}